\documentclass[runningheads]{llncs}
\usepackage[T1]{fontenc}
\usepackage{graphicx}
\usepackage{fontawesome}
\usepackage{todonotes}
\usepackage{hyperref}
\hypersetup{breaklinks=true} 
\usepackage{svg}
\begin{document}
\title{Defining, Understanding, and Detecting \\Online Toxicity: Challenges and Machine Learning Approaches}
\titlerunning{Online Toxicity}

\author{Gautam Kishore Shahi\inst{1}\orcidID{0000-0001-6168-0132} \and\\
Tim A. Majchrzak\inst{2,3}\orcidID{0000-0003-2581-9285} }
\authorrunning{Shahi \& Majchrzak}

\institute{University of Duisburg-Essen, Germany
\and Center for Advanced Internet Studies (CAIS), Germany
\and
University of Bochum, Germany\\
\email{gautam.shahi@uni-due.de}
}
\maketitle              
\begin{abstract}
Online toxic content has grown into a pervasive phenomenon, intensifying during times of crisis, elections, and social unrest. A significant amount of research has been focused on detecting or analyzing toxic content using machine-learning approaches. The proliferation of toxic content across digital platforms has spurred extensive research into automated detection mechanisms, primarily driven by advances in machine learning and natural language processing. Overall, the present study represents the synthesis of 140 publications on different types of toxic content on digital platforms. We present a comprehensive overview of the datasets used in previous studies focusing on definitions, data sources, challenges, and machine learning approaches employed in detecting online toxicity, such as hate speech, offensive language, and harmful discourse. The dataset encompasses content in 32 languages, covering topics such as elections, spontaneous events, and crises. We examine the possibility of using existing cross-platform data to improve the performance of classification models. We present the recommendations and guidelines for new research on online toxic consent and the use of content moderation for mitigation. Finally, we present some practical guidelines to mitigate toxic content from online platforms.

 \and Datasets \and Content Moderation

\end{abstract}

\section{Introduction}
\label{sec:1}
With the emergence of Web 2.0, which enables users to produce and consume content on digital platforms such as social media and news media, the Internet has witnessed significant growth in accessibility and user engagement~\cite{o2007web}. Simultaneously, the global internet user base expanded rapidly, currently covering more than 68\% of the world population.\footnote{\url{https://www.itu.int/itu-d/reports/statistics/2024/11/10/ff24-internet-use/}} While user-generated content proliferated, the dissemination of toxic content -- such as hate speech and cyberbullying -- also increased~\cite{fulantelli2022cyberbullying}. Subsequently, with the intervention of government regulations (such as the European Union (EU) Digital Service Act (DSA) ~\cite{shahi2025year}, United Kingdom (UK)'s Online Safety Act~\cite{uk_online_safety_2023}) and the enforcement of content moderation policies by social media platforms, efforts were made to regulate and mitigate such content. However, despite these measures, the creation and dissemination of toxic content persist.

Toxic content appears in multiple forms, such as hate speech and offensive speech and in different data formats~\cite{wulczyn2017ex}. One of the commonly occurring toxic content is hate speech, which has become a widespread problem~\cite{jahan2023systematic}. With easy access to social media platforms, such as Twitter (now \emph{X}), YouTube, and Gab, the amount of toxic content has been increasing~\cite{shahi2022mitigating}  over the years. The topic of toxic content is linked to global developments and recent crises, such as hate speech on the Russo-Ukrainian conflict~\cite{di2023hate}, COVID-19~\cite{shahi2022mitigating}, Israel-Hamas Conflict~\cite{shahi2024fakeclaim} and elections in different countries~\cite{shahi2024agenda}. Digital platforms enable users to share content in various media formats and languages while requiring them to adhere to platform-specific guidelines. For example, on YouTube, users can post-attack someone or a group by posting videos or commenting on existing videos in multiple languages. 

Prior research shows that the majority of studies were done with specific languages, most notably English, and on specific platforms such as Twitter~\cite{siegel2020online}, Facebook~\cite{del2017hate}, and YouTube~\cite{doring2020gendered}. However, the presence of toxic content is not limited to one platform or language. In the last couple of years, multiple platforms such as TikTok, Reddit, and Bluesky have grown a considerable user base but have not been researched well. Toxic content has been studied using various approaches, including machine learning-based classification and user studies. In particular, researchers have developed models to detect toxic content, which require large-scale, fine-grained annotated datasets for effective training~\cite{garg2023handling}.
Gathering such high-quality training data is expensive and time-consuming~\cite{shahi2022amused}. Thus, Every research endeavour needs to start from scratch, i.e., data collection, annotation, and building a classification model, which is time-consuming and practically challenging.

Subsequently, a classification model was developed to detect hate speech in YouTube comments, demonstrating that augmenting training data with additional datasets based on definitions and content similarity can significantly enhance model performance~\cite{webist24}. Previous research shared different datasets without providing insightful information such as definitions used and annotation guidelines~\cite{poletto2021resources,al2019detection,fortuna2018survey}. However, datasets vary considerably by social media platform, period of data collection, and linguistic style~\cite{al2019detection}. Hence, in prior research, it is difficult to identify the similarity of datasets across different platforms.  This motivated us to conduct a literature review of existing literature on toxic content, current guidelines and regulations, and the platform's policy and provide a comprehensive understanding based on the topic of research, definitions, and annotation goodliness. It will be helpful in selecting an appropriate dataset as an additional training dataset for the classification model.  

Having a unified platform that provides a descriptive view of hate speech will help to do cross-platform studies and cross-topic studies, and to use of an existing dataset for building a classification model. The key contributions of our research are:
\begin{itemize}
\item Provide a comprehensive overview of existing research on toxic content across digital platforms.
\item Present a detailed analysis of datasets and highlight their key parameters, such as annotation guidelines, definitions of hate speech, and the proportion of toxic content.
\item Offer a publicly accessible repository to promote data sharing and enhance dataset reusability.
\end{itemize}

The current study is an extended version of conference paper \cite{iceis25}, which explains providing additional data helps in performance of prediction model. All datasets used in the study are provided at a GitHub repository.\footnote{\url{https://github.com/Gautamshahi/ToxicContent}}

This article is organized as follows: Section~\ref{sec:2} discusses the related work on toxic content, followed by the research method in Section~\ref{sec:3}. We then show our experiment and results in Section~\ref{sec:4} before discussing the findings in Section~\ref{sec:5}. Finally, we discuss the ideas for future work in Section~\ref{sec:6}. 

\section{Related Work and Background}
\label{sec:2}
Toxic content has been studied by experts on different subjects for its detection, patterns, and impact on society. 
Machine learning is the dominant approach to classification in various domains such as political sentiment analysis~\cite{rochert2020opinion}, detecting incivility and impoliteness in online discussions~\cite{stoll2020detecting} as well as the classification of political tweets~\cite{charalampakis2016comparison}. Especially the state-of-the-art technique BERT (Bidirectional Transformers for Language Understanding)~\cite{malmasi2017detecting} has been used for the detection of toxic contents~\cite{zhang2024efficient,salminen2020developing,aggarwal2019ltl,liu2019nuli,zampieri2019semeval}. Prior research has done predominately on text data, and text classifier aims to train a robust classifier to recognize hate comments globally, i.e., on different platforms and languages. For the identification and training of the models, several studies classify hate speech using deep neural network architectures or standard machine learning algorithms. Wei et al. compare machine learning models on different public datasets~\cite{wei2017convolution}. The results indicate that their approach of a convolution neural network outperforms the previous state-of-the-art models in most cases~\cite{wei2017convolution}. A recent study used transfer learning~\cite{yuan2023transfer} to train 37,520 English tweets, showing a trend towards more complex models and better results.
Prior studies have analyzed hate speech on YouTube; one study highlights the hate speech on YouTube on Syrian refugees~\cite{aslan2017online}.
In another study, hate speech on gender is studied on a small dataset quantitatively~\cite{doring2019fail}. Vidgen \& Derczynskivid provides a literature review on abusive languages by highlighting the issue in creating training datasets and possible direction for combining more datasets for helping data-driven research~\cite{vidgen2020directions}.

Previous studies have mainly focused on detecting toxic content in different languages~\cite{ousidhoum2019multilingual} and on comparing different social media platforms~\cite{salminen2020developing}. Considering the past studies about toxic content, it is noticeable that many studies only concentrate on one platform or specific language, such as English~\cite{waseem2016you}, German~\cite{ross2016hatespeech}, Spanish~\cite{ben2016hate} and Italian~\cite{del2017hate} to train a machine learning model to automatically classify and predict which unseen texts can be considered hate speech.
Ousidhoum et al. applied a multilingual and multitasking approach to train a classifier on three languages (English, French, and Arabic) based on Twitter tweets using a comparison of traditional machine learning and deep learning models~\cite{ousidhoum2019multilingual}. In another work, a Twitter-like user interface was built to flag hate speech and it helps in user awareness for hateful content~\cite{10.1007/978-3-031-19097-1_34}. 

Previous research uses annotated datasets for the supervised classification model. Beyond the different methods for training the model, different guidelines are used for data annotation. At the same time, \cite{davidson2017automated,waseem2016hateful} opted for a multi-label procedure (\emph{hateful}, \emph{offensive (but not hateful)}, and \emph{neither}~\cite{davidson2017automated}; \emph{racist}, \emph{sexist}~\cite{waseem2016hateful}), the annotation of the data of~\cite{ross2016hatespeech} was collected with a binary labeling schema (hate speech as \emph{yes} or \emph{no}). 

However, research is needed to consider the classifiers' performance in multilingual contexts in combination with datasets from social media platforms. Salminen et al.~\cite{salminen2020developing} points out that the mono-platform focus in hate speech detection research is problematic, as there are no guarantees that the models developed by researchers will generalize well across platforms. This issue underscores the importance of our research in exploring cross-platform generalization.  Fortuna et al. have shown that merging and combining different datasets can enhance the overall performance of classification models~\cite{fortuna2018survey}. However, a more systematic evaluation is needed to determine the extent of improvement when a classifier is trained on one dataset, and then another is added. This approach could potentially lead to significant advancements in hate speech detection. The role of LLMs such as ChatGPT in generated annotated data without any background truth is still under exploration~\cite{li2023hot}. Consequently, toxic content detection still depends on gathering human-annotated data to build a classification model.

In the prior work it has been shown that a dataset with similar content and definition helps in improving the performance of a classification model. However, there is a lack of resources which provide the description of hateful words. Datasets on toxic content are created on different platforms, in different languages and with different topics of hate; having a unified platform which provides a descriptive view of toxic content will be helpful for doing a cross-platform study or cross-topic study or for using existing datasets for building classification models. Hence, we conducted a literature review of prior research on toxic content.

\section{Research Method}
\label{sec:3}
This section explains the process of collecting publications using the PRISMA framework~\cite{page2021prisma}. The eligibility criteria to research work is that they should discuss new datasets for any kind of toxic content regarding people or groups of people, such as hate speech, cyberbullying, interpersonal harassment, toxic language, trolling, aggression, bullying and profanity. The search was covered different languages, topic of datasets, platforms (such as social media, online news source), definition provided, and data extraction processes. We have not included the datasets for countering toxic content or explanation of toxic content because the goal is to have more similar data to enhance the performance of the classification model. 

\begin{figure}[htp]
    \centering
    \includegraphics[width=0.96\textwidth]{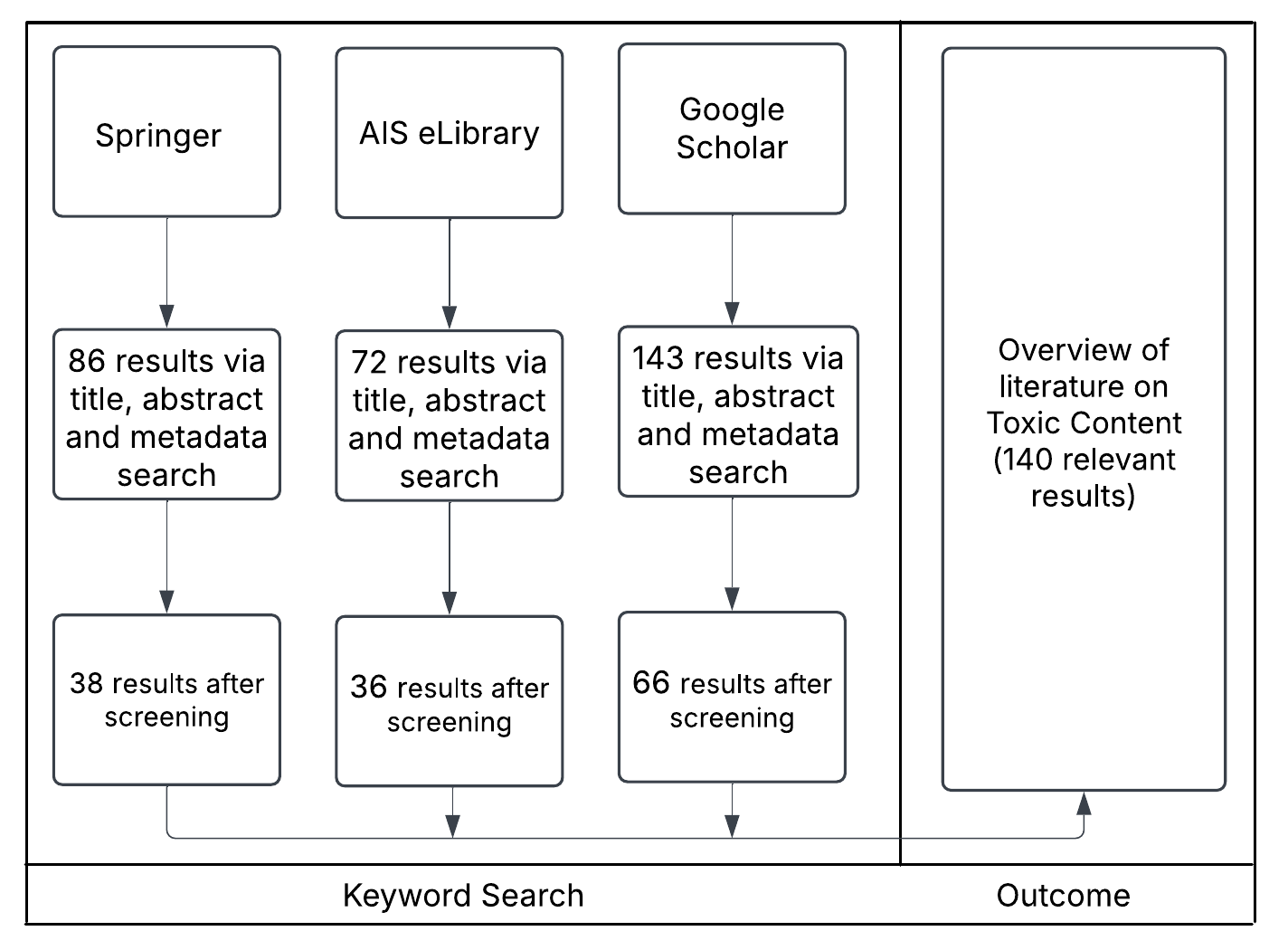}
    \caption{Overview of Publications During the Literature Search}
    \label{fig:ResultsLiterature}
\end{figure}

\subsection{Sources of Content} 
Toxic content was discovered from a wide range of bibliography sources as described below:
\begin{itemize}
    \item Google Scholar is a freely accessible search engine that indexes scholarly literature across various disciplines, including journal articles, conference papers, theses, books, and patents. It covers publication from different sources. 
\item Springer: Springer offers a wide collection of research articles. Springer covers many conference proceedings.
\item AISEL: A central repository for research papers and journals published in the information system community.
\end{itemize}

\subsection{Search}
In the initial search, Google Scholar returned 143 results, of which 66 were deemed relevant to this study.
AISEL yielded 72 results, 36 of which met our criteria, while Springer provided 86 results, 38 of which were considered appropriate. Two researchers independently screened the titles and abstracts to identify suitable articles. In cases of disagreement, the full text of the article was reviewed collaboratively, and consensus was reached through discussion. Overall, we have got 140 publications for this literature review. An Overview of search process and number of publications obtained is shown in Figure \ref{fig:ResultsLiterature}.

\begin{table}[tp]
\caption{Description of Metadata of Collected Datasets}
\begin{tabular}{p{2.5cm}p{6.5cm}p{3cm}}
\hline
Element &  Description & Example\\
\hline
Dataset Number  & Alphanumeric number to uniquely identify the dataset & HS1 \\
Year & Year in which the research is published & 2022 \\
Dataset name & A name given to the dataset (optional) & Hateful Memes Dataset\\
Platform & Name of the social media platform from where the data is collected &  Facebook\\
Language & Language of content in data & English\\
Dataset available & Check in the paper or do a google search to find the dataset & Publically Available\\ 
Dataset available at & If the link is available or obtained from google search then mention the link of the dataset & \href{https://ai.meta.com/tools/hatefulmemes/}{(Link)}$^*$ \\
Data format & Multi-modality of data & Image+text \\
Evaluated in & Time used to crawl the data & 23-04-2020 to 23-06-2020 \\
Search Keywords & Name or list of hashtags or keywords used in the paper to gather the data & -- \\
Definition & If a definition of toxic content is provided in the paper & Language that expresses prejudice against a person from a group or a group, primarily based on ethnicity, religion, or sexual orientation. \\
Data Crawling Process & How was the data crawled & own tool or python/java library \\
Topics covered & Topic od analysis of toxic content &  racism, sexism, refugee \\
Size of dataset & Size of data of the crawled dataset that has been annotated/coded & 24\,582 tweets \\
Annotation Guidelines & If annotation guidelines is available. & -- \\
Annotation Process & Process of annotating toxic content & Aamazon Mechanical Turk(AMT) \\
Number of Annotators & Number of annotator used & 2 \\
Intercoder measure &  Measure to show intercoder agreement & Cohen's kappa \\
Classification model used & name of classification model used &  BERT, LSTM \\
\hline
\scriptsize
&* https://ai.meta.com/tools/hatefulmemes/ &
\end{tabular}
\label{data:metadata}
\end{table}

\section{Description of Datasets}
\label{sec:4}
Different metadata elements have been identified for analyzing each publication’s dataset, as detailed in the Table \ref{data:metadata}. 
First, we defined a unique identifier for the date, then followed with the dataset name (if given by authors), platforms covered, languages, data collection and annotation process, the definition of toxic content and the research method used. Prior research~\cite{webist24} suggests that for datasets having similar definitions, content, and hate words,  a combination of these datasets helps to improve the classification performance. In the following, the elaborate on the criteria.

\subsection{Language} The collected dataset are dominated by English content~\cite{grigore2011increasing,gitari2015lexicon}, which appears over a hundred times. Other frequently researched language are Arabic~\cite{mubarak2017abusive,ousidhoum2019multilingual}, Hindi~\cite{kumaraggression,mandl2020overview}, German~\cite{ross2016hatespeech,webist24}, Indonesian~\cite{ibrohim2019multi,8355039}, Russian~\cite{zueva2020reducing,bodrunova2015twitter} and Spanish~\cite{giglietto2015or,charitidis2020towards}, each appearing multiple times. French~\cite{charitidis2020towards}, Turkish~\cite{arcan2013interrupted} and, Italian~\cite{SanguinettiEtAlLREC2018} are also present.

Languages such as Greek~\cite{charitidis2020towards}, Urdu~\cite{rizwan2020hate}, Bengali~\cite{romim2021hate,al2024hate}, Vietnamese~\cite{luu2020comparison}, Czech~\cite{hrdina2016identity}, Croatian~\cite{ljubevsic2018datasets}, Polish~\cite{ptaszynski2019results}, Danish~\cite{ajvazi2022dataset}, Korean~\cite{lee2022k}, and Hingish~\cite{mathur2018did,bohra2018dataset} each appear only a couple of times. 
Some dataset contain multiple languages. The reason for content having in multiple languages could be the dominance of major digital platforms in English language and, presence of an extensive tools and library for Natural Language Processing (NLP) processing such as Natural Language Toolkit (NLTK)\footnote{\url{https://www.nltk.org/}}, spaCy\footnote{\url{https://spacy.io/}}.

\subsection{Platforms}  
Most toxic content studies collected their data from social media platforms, with Twitter serving as the primary source~\cite{10.1007/978-3-031-19097-1_34}, mainly due to ease of access of dataset (prior to becoming \emph{X}). Other frequently mentioned platforms include Facebook~\cite{kumaraggression,hrdina2016identity}, YouTube~\cite{webist24,shammur2020offensive} and Reddit~\cite{gibson2017safe}. A few research have been done on the content from  Instagram~\cite{putra2020hate}, Gab~\cite{kennedy2022introducing}, Whisper~\cite{silva2016main} and Yik Yak~\cite{7235779}. In terms of news media~\cite{benvcek2016refugees}, research has been done on the content from Al Jazeera, Fox News~\cite{gao2017detecting}, SacBee.com~\cite{diakopoulos2011towards}, and several Turkish newspapers~\cite{arcan2013interrupted}. Notably, some data come from synthetic sources, radio shows~\cite{josey2010hate}, gaming platforms (like World of Warcraft and League of Legends~\cite{bretschneider2017detecting}), and forums such as Stormfront~\cite{gitari2015lexicon}. Some research also targets community platform such as Wikipedia~\cite{wulczyn2017ex,oak2019poster,9006336,grigore2011increasing}. Overall, datasets incorporate a diverse range of platforms but are clearly dominated by social media, reflecting its central role in online discourse and toxic content analysis.

For data collection, a wide range of approaches were used from different platforms. For social media, the researcher used APIs to collect data using keywords~\cite{basile2019semeval,kwok2013locate,mandl2020overview,qian-etal-2019-benchmark} and accounts~\cite{yardi2010}. Researchers typically used the official API to gather data from Twitter, Facebook, YouTube, Reddit, and Whisper. Public data dumps like Pushshift.io were also utilized to gather data from  Wikipedia talk pages\footnote{\url{https://en.wikipedia.org/wiki/Help:Talk_pages}} and Gab posts. Twitter data was typiclaly collected from API\footnote{\url{https://developer.x.com/en/docs/x-api}} using keywords, hashtags, or geo-locations. Several studies employed tools like Sifter (in partnership with Gnip), DiscoverText, or custom-built Twitter engines~\cite{stieglitz2022smart} to facilitate efficient data retrieval. Other research used web scraping from comment sections of news websites, crawling blogs from hate directories, and manual capture of platforms like Yik Yak~\cite{7235779} via screenshots. Some studies also collected data \emph{offline} through simulation and role-play exercises involving participants~\cite{sprugnoli2018creating}. Once data was collected, researchers used the sampling method for data annotation, and to ensure balanced sampling, some studies adopted random walk algorithms and days-based sampling. 

Overall, a combination of automated APIs, keyword-based crawling, manual methods, and existing datasets were used to compile robust corpora for analyzing online discourse and toxic content.

\subsection{Topic Covered}  For the analysis of toxic content, researchers have focused on different topics for data collections. Researchers covered content related on evolving topics such as refugees~\cite{zhang2018hate}, immigrants~\cite{berglind2019levels}, and migrants~\cite{ousidhoum2019multilingual}. In terms of personal attacks, toxic content was analyzed on topics such as LGBTQ+ phobia~\cite{leite-etal-2020-toxic}, misogyny~\cite{bhattacharya-etal-2020-developing}, and harassment toward individuals with Autism Spectrum Disorder (ASD)~\cite{davidson2017automated}. Researchers covered specific events such as the George Tiller shooting~\cite{yardi2010}, the Charlie Hebdo attack~\cite{bretschneider2014detecting}, and the murder on a Romani family in Olaszliszka~\cite{vidra2012rise}. 
Some studies addressed the dark side of Web such as vulgar and pornographic content~\cite{mubarak-etal-2017-abusive} and obscene speech~\cite{leite-etal-2020-toxic}, often in connection with online harassment or personal attacks. Overall, the topics reflect the varied and evolving nature of hate speech and offensive communication in online spaces which are summarized below.

\begin{itemize}
    \item \textbf{Identity-Based Hate and Discrimination}
    \begin{itemize}
        \item Ethnicity: Racism, anti-Hungarian attacks, race-based hate speech, ethnic discrimination.
        \item Religion: Islamophobia, anti-Muslim sentiment, religious intolerance, attacks related to religious identity (e.g., Charlie Hebdo shooting).
        \item Gender and Sexism: Misogyny, sexism, feminism-related hate, appearance-based and intellectual/political bias against women.
        \item Sexual Orientation and LGBTQ+: Hate towards LGBTQ+ communities, sexuality-related hate, body shaming.
        \item Disability: Hate towards individuals with disabilities (e.g., ASD), discrimination based on handicap or health condition.
        \item Immigrants and Refugees: Hate toward immigrants, migrants, and refugees; xenophobia; issues around migration policies.
    \end{itemize}

    \item \textbf{Sociopolitical Conflicts and Controversial Events}
    \begin{itemize}
        \item Violence and Crimes: Murders in Olaszliszka and Tatárszentgyörgy, George Tiller shooting.
        \item Geopolitical Conflicts: Israel-Palestinian conflict, attacks on refugees.
        \item Political and Ideological Tensions: Political correctness, Islamic leftism, feminist debates, US elections.
        \item Anti-fascist Protests: Communication and coordination during protests.
    \end{itemize}

    \item \textbf{Media and Platform-Related Issues}
    \begin{itemize}
        \item Online Content and Platform Abuse: Malicious Facebook pages, phishing, spam, untrustworthy information, misleading claims.
        \item Toxicity and Health Judgements: Evaluations of online content based on perceived toxicity or emotional harm.
        \item Controversial Media: Celebrity footage, provocative or offensive media content.
    \end{itemize}

    \item \textbf{Mixed and Unspecified Themes}
    \begin{itemize}
        \item No Specific Topic: General hate speech without specific targets or categories.
        \item Multiple Combined Themes: Entries covering combined aspects (e.g., race, religion, gender).
        \item Other Keywords: Knowledge reuse, vulgar speech, emotional responses, controversial humor.
    \end{itemize}
\end{itemize}

\subsection{Definition of Toxic Content}
Before proceeding to studying toxic content, the definition of toxic content is important in terms of the development of a code book and performing data annotation. There is a lack of common understanding definition of different toxic content~\cite{webist24}. Each toxic content is defined differently for data annotation; we looked at different sources for providing definitions, such as UNICEF\footnote{\url{https://www.unicef.org/end-violence/how-to-stop-cyberbullying}} defines cyberbullying as spreading lies or sharing information which harms, abuse or threatens others users. It can happen online or offline, but is most commonly seen on social media. In 2021, the European Union (EU) formed a law to counter hate crime and hate speech. Peršak defines hate speech as a direct attack on people based on what we call \emph{protected characteristics}: race, ethnicity, national origin, religious affiliation, sexual orientation, caste, sex, gender, gender identity, and serious disease or disability~\cite{persak2022criminalising}. Nockleby defines hate speech as any communication that disparages a person or a group based on some characteristics such as race, color, ethnicity, gender, sexual orientation, nationality, religion, or other characteristics~\cite{nockleby2000hate}. 

However, over the years, the coverage and granularity of toxic content has changed. Both governments and platforms come up with extensive definitions of toxic content. With the implementation of several government regulations such as EU DSA, the UK's Online Safety Act 2023\footnote{\url{https://www.legislation.gov.uk/ukpga/2023/50}}, and USA's Kids Online Safety Act~\cite{alemi2023support}. Platforms have adapted their policies and laws in dealing with toxic content. Here, we are listing the platform guidelines provided by very large online platforms (VLOPs) social media platforms under EU DSA. Platforms use both manual and automated methods to detect and remove content. Platforms also allow users to flag or report suspicious content, which is later moderated. 

\textbf{Meta}, the parent company of Facebook, Instagram, and WhatsApp, started defining toxic content as bullying and harassment in 2019 and continuously updates this definition regularly~\cite{meta_bullying}. Presently, Meta defines bullying and harassment for public figures and private individuals, giving higher protection to private individuals. Content intended to degrade or shame may be removed or demoted. \textbf{Google} (or rather Alphabet), a parent company of YouTube, provides community guidelines for \textit{Violent or dangerous content}, which includes  Harmful or dangerous content and Hate speech, and content falling in this category is not allowed on the platform~\cite{youtube_harassment}. \textbf{X (formerly Twitter)} covers toxic content as abuse and harassment, which include different types of offensive content that is not allowed on the platform. It has removed posts, suspended accounts, and made posts less visible~\cite{x_abusive_behavior}. \textbf{TikTok}, a platform platform for video sharing, defines community guidelines for abuse and harassment content and removes and restricts content on the platforms~\cite{tiktok_community_guidelines}. \textbf{LinkedIn} prohibits bullying or harassment, including abusive language, unwanted advances, and encouraging others to harass a member on LinkedIn, and these contents are removed once detected~\cite{linkedin_ads_settings}. \textbf{Snapchat} provides community guidelines to monitor \textit{Sexual content, thereat, violence, harm, hateful content, harassment and bullying}~\cite{snapchat_community_guidelines}. \textbf{Pinterest} includes harmful, hateful, or violent content or behavior in the community guidelines~\cite{pinterest_community_guidelines}. Pinterest may take action by removing, restricting, or limiting the reach of such content -- as well as the accounts, individuals, groups, or domains responsible for creating or spreading it -- depending on the level of harm it presents.

\subsection{Annotation Technique} For building classification models for toxic content, often research collected raw data from online platforms. Annotation processes were performed by using a mix of in-house and crowdsourcing methods~\cite{zhang2018hate,9006336}. In-house annotation~\cite{davidson2017automated,8355039} includes data annotation using humans as well as automated tools. The common practice of in-house annotation involves researchers, student assistants, volunteers, and native speakers. Crowdsourcing annotation is performed using third-party software or websites; The latter hire annotators from different backgrounds. Crowdsourcing annotation is rather fast, expensive, and commonly offers quality checks. Some of the crowdsourcing platforms widely used are CrowdFlower~\cite{mubarak2017abusive} and Mechanical Turk~\cite{qian-etal-2019-benchmark}. Crowdsourcirng annotation is done either independently or in conjunction with in-house annotation to ensure quality and scalability. A few studies adopted a mixed approach, starting with in-house annotation and then validated the results using crowdworkers. However, not all studies specified their annotation procedures.

\subsection{Models Used}
Once data is collected and annotation is completed, several approaches have been applied for achieving a deeper understanding.
For toxic content as a classification problem,  traditional models include traditional machine learning classifiers such as Naive Bayes (NB)~\cite{davidson2017automated,jigsaw2018toxic}, Logistic Regression (LR)~\cite{albadi2018they}, Support Vector Machines (SVM)~\cite{jha2017does,de-gibert-etal-2018-hate,bohra2018dataset,fersini2018overview}, and Random Forests (RF)~\cite{8355039}. Other researchers used Liblinear \& Multinomial Naive Bayes (MNB)~\cite{6406313}, and Stochastic Gradient Descent Classifier (SGDC)~\cite{pitenis2020offensive} for binary and multi-class classification problems~\cite{mulki2019hsab}. In terms of features, these models use textual features such as n-grams, TF-IDF~\cite{bretschneider2017detektion}, and lexicon-based features.

With the development of deep learning, complex models such as Convolutional Neural Networks (CNNs)~\cite{zampieri2019semeval,baziotis-pelekis-doulkeridis:2017:SemEval2}, Recurrent Neural Networks (RNNs)~\cite{qian-etal-2019-benchmark}, Long Short-Term Memory (LSTM)~\cite{zhang2018hate}, bi-LSTM~\cite{gao2017detecting} with attention, Gated Recurrent Unit (GRU) \cite{dey2017gate}, and CNN+GRU hybrids have gained popularity and used for classification.
Several researchers used ensemble methods, such as combining CNN with GRU, LSTM, and attention-based models, have been deployed to capture contextual and sequential nuances in language. Transformer-based models such as BERT~\cite{webist24}, M-BERT~\cite{leite-etal-2020-toxic}, ULMFiT~\cite{gertner2019mitre}, and Bag of Word(BoW) with AutoML~\cite{leite-etal-2020-toxic} enhanced models are also widely used for transfer learning and zero-shot scenarios.
Some researchers have applied Graph-based models such as GraphSAGE~\cite{ribeiro2018characterizing}, as well as embedding modeling~\cite{djuric2015hate} paragraph2vec and CBOW for capturing feature for classification models. Additionally, some studies used Linguistic Inquiry and Word Count (LIWC)~\cite{bretschneider2017detecting}, SentiStrength~\cite{grigore2011increasing}, and sentiment analysis~\cite{grigore2015impact} for analyzing psycholinguistics approaches~\cite{6406313} of toxic content.

\section{Discussion}
\label{sec:5}
In previous studies, researchers focused on a variety of topics. Toxic content is not limited to online platforms; it may lead to violence and social unrest, including but not limited to public protest~\cite{10.1007/978-3-031-71210-4_7} and even to  incidents such as mob lynching~\cite{10.1007/978-3-031-43590-4_7}. Most research focused on Twitter and Facebook; however, collecting data from these platforms is no longer feasible with free APIs \cite{toolittle}. 
In terms of language, research was dominantly performed in English. However, each country has a public discourse of its own, leading to a variety of languages that need to be studied. 

For studying toxic content, collecting datasets is a vital step, which requires choosing the proper platforms and keywords to select relevant posts. For instance, YouTube allows to scrap YouTube videos and comments; however, keywords search leads to a lot of noise with a wide range of videos that do not fit with the criteria. Hence, a manual check of videos is required.  
Furthermore, based on data requirements adding a dataset from existing research will be helpful in adding a training dataset for classification as proposed by~\cite{webist24}. Data dumps are also available on PushShift and through the Internet Archive. However, such datasets might be only interesting for analyzing user behavior or training data because of the evolving patterns of platforms and content, but the latest datasets are always useful.

For tackling the problem of toxic content, the research community actively organizes workshops and challenges to gather evolving approaches to counter toxic content. Organizers collect data for toxic content and ask participants to make a classification model for the detection of toxic content. One of the events is HASOC~\cite{mandl2019overview}, which has organized competitions for hate speech, offensive speech, and counter hate speech detection in Hindi, English, and German since 2019. Similar workshops and challenges, such as EVALITA~\cite{bosco2018overview}, SEMEVAL~\cite{basile2019semeval}, CONSTRAINTS~\cite{sharma2022constraint}, and PAN@CLEF~\cite{rangel:2021} focus on different aspects of toxic content, including hate speech, offensive content, and online abuse.

Not only research communities but also companies organizes competition on this topic; the Facebook memes challenge~\cite{kiela2020hateful} was organized with artificial datasets of hateful memes. Jigsaw organized a challenge~\cite{jigsaw2018toxic} to detect toxic content using Wikipedia content. These competitions played a crucial role in advancing the field of toxic content detection. Also, they provide crucial datasets to explore research such as the Facebook memes datasets explored by researchers from natural language processing, computer vision, and sociology to analyze computation and human factors by consuming toxic contents. However, the participants fail to make a generalizable and universal model for the detection of toxic contents.
With the proliferation of generative AI, particularly Large Language Models (LLMs) have been used for different tasks such as data annotation~\cite{iceis25} and data profiling. In the future, data annotation can be performed by LLM, which might reduce cost and time in preparing datasets for studying toxic content. Also, interpretable machine learning models are able to explain results~\cite{nandini2022explaining,iceis25} to make detection of toxic content transparent.  

With the emergence of acts and regulations from different countries aksed platforms for content moderation. In the last couple of years, platforms have extensively provided guidelines for the allowed content, and they regularly moderate to mitigate the toxic content. Platforms use both manual and automatic approaches for identifying content. Also, the EU provides an option to demand datasets for research, which might be an interesting way of collecting moderated content from platforms and use for study.  This will also help remove the cost of data collection and annotation for studying toxic content.

It needs to be mentioned as an ethical outlook that the same tools and techniques that can be used to remove toxic content could be exploited to \emph{moderate} content that by some criterion is considered undesirable. While there is a understanding in the research community what facilitates toxic content, countries might declare certain opinions, standpoints, ideologies, or even scientific  consensuses or facts to be \emph{toxic} and make them, thus, subject to removal. This mandates particular care for the research described in this article, as well as it mandates careful weighing of freedom of speech versus content moderation, ultimately aiming at a pleasant, non-harmful experience for all people being online.

\section{Conclusion \& Future Work}
\label{sec:6}
The rise of toxic content, especially hate speech, across digital platforms has prompted a surge in research focused on its detection and mitigation. This literature review presents findings from prior studies, emphasizing the definitions, data sources, annotation guidelines, and machine learning approaches used in hate speech detection. Findings show that the present studies are mainly focused on English and mono-platform sources. Despite much research, hate speech detection in under-resourced languages, particularly regional languages, is limited. Our review shows that cross-platform and cross-lingual datasets can significantly enhance the performance and robustness of classification models. As a step toward this, we propose practical guidelines for researchers and practitioners planning to create or curate new datasets in this domain. 

With the changes in platform policy and monetary demand, data collection from social media platforms such as Facebook, Instagram, and X is challenging, which limits the data collection. However, having limited datasets or dataset from less research platforms such as YouTube, Reddit we might use the 
existing datasets based on definitions and topics to increase the training size of dataset for machine learning models. 

With the emergence of government guidelines such as the EU DSA, platforms are actively monitoring contents. They have extensive guidelines for moderation of toxic content in the platforms, which might help in mitigating toxic content. 

Overall, our work serves as a foundation for future research and development efforts aimed at mitigating online toxicity in a more holistic, inclusive, and scalable manner.

\section*{Data Sharing}
\label{sec:7}

The dataset collected in the present study will be provided on GitHub\footnote{\url{https://github.com/Gautamshahi/ToxicContent}}, allowing researchers to use it based on different criteria. GitHub repository includes detailed annotations and metadata, enabling reproducibility and further analysis across various research contexts.

\bibliographystyle{splncs04}
\bibliography{example}

\end{document}